# Adaptive Transformer Attention and Multi-Scale Fusion for Spine 3D Segmentation

Yanlin Xiang*
University of Houston
Houston, USA

Qingyuan He
New York University
New York, USA

Ting Xu
University of Massachusetts Boston
Boston, USA

Ran Hao
University of North Carolina at Chapel Hill
Chapel Hill, USA

Jiacheng Hu
Tulane University
New Orleans, USA

Hanchao Zhang
New York University
New York, USA

*Abstract*-This study proposes a 3D semantic segmentation method for the spine based on the improved SwinUNETR to improve segmentation accuracy and robustness. Aiming at the complex anatomical structure of spinal images, this paper introduces a multi-scale fusion mechanism to enhance the feature extraction capability by using information of different scales, thereby improving the recognition accuracy of the model for the target area. In addition, the introduction of the adaptive attention mechanism enables the model to dynamically adjust the attention to the key area, thereby optimizing the boundary segmentation effect. The experimental results show that compared with 3D CNN, 3D U-Net, and 3D U-Net + Transformer, the model of this study has achieved significant improvements in mIoU, mDice and mAcc indicators, and has better segmentation performance. The ablation experiment further verifies the effectiveness of the proposed improved method, proving that multi-scale fusion and adaptive attention mechanism have a positive effect on the segmentation task. Through the visualization analysis of the inference results, the model can better restore the real anatomical structure of the spinal image. Future research can further optimize the Transformer structure and expand the data scale to improve the generalization ability of the model. This study provides an efficient solution for the task of medical image segmentation, which is of great significance to intelligent medical image analysis.

## I. Introduction

Spine 3D semantic segmentation is a critical task in medical image processing, especially in the diagnosis and treatment of spinal diseases. With the advancement of medical imaging technologies, particularly the widespread use of CT and MRI imaging techniques, 3D reconstruction of the spine has become an indispensable component of diagnostic systems. However, due to the complexity of the spine's anatomical structure and various factors such as patient body type, posture, and disease variations, traditional manual analysis methods face significant challenges in the process of spinal image segmentation. Addressing these challenges, how to leverage deep learning algorithms to improve the accuracy and efficiency of spine 3D semantic segmentation has become one of the key research hotspots in the field of medical image processing [1,2].

In recent years, deep learning, especially Convolutional Neural Networks (CNNs), has achieved remarkable progress in medical image analysis. Traditional 2D convolutional networks are limited in handling spatial information in 3D medical images, which has led to the development of 3D Convolutional Neural Networks (3D-CNNs). While 3D-CNNs can better capture 3D spatial information, the training process of deep networks still faces challenges such as high computational complexity and large data requirements. To address these issues, Transformer-based models have gained attention in recent years. The Swin Transformer, as an efficient image processing architecture, has shown excellent performance and has become a promising choice for medical image processing. The SwinUNETR model, which combines the Swin Transformer with U-Net, has demonstrated remarkable performance in medical image segmentation tasks, especially in dealing with complex structures and fuzzy boundaries, outperforming traditional methods.

However, applying SwinUNETR to spine 3D segmentation presents challenges due to the spine's intricate structure, individual anatomical variations, and the presence of subtle pathological features. Existing models may struggle with capturing fine-grained details and preserving critical boundary information. Therefore, further improving SwinUNETR to better address these specific challenges is the primary goal of this study.

This research proposes an enhanced SwinUNETR model tailored to spinal image characteristics, incorporating advanced attention mechanisms and multi-scale feature fusion to improve detail preservation, boundary clarity, and segmentation accuracy [3]. By improving segmentation precision, the proposed method not only supports more accurate spinal disease diagnosis and treatment planning but also contributes to advancing intelligent medical diagnosis and promoting the development of precision medicine. The proposed methods may also benefit other medical image segmentation tasks, contributing to broader applications of deep learning in healthcare [4].

## II. RELATED WORK

The advancement of deep learning techniques has provided a powerful foundation for addressing complex segmentation tasks, particularly in volumetric medical imaging. Notably, research into optimized U-Net variants equipped with attention modules has demonstrated how the integration of multi-scale feature fusion and adaptive attention enhances spatial consistency across multiple resolutions, offering strategies that directly contribute to the design of the enhanced segmentation model proposed in this paper [5].

Incorporating graph neural networks (GNNs) has further strengthened the ability of segmentation models to capture spatial dependencies and structural relationships between anatomical components. GNN-based hierarchical mining approaches have shown particular effectiveness in representing complex spatial arrangements and feature hierarchies, providing valuable techniques for refining spatially-aware feature fusion in 3D segmentation tasks [6]. The evolution of Transformer-based architectures, especially those enhanced with multi-level attention mechanisms, offers additional strategies for improving the adaptability and precision of feature extraction processes in segmentation models [7]. Similarly, adaptive attention embedding techniques, which dynamically reweight features during encoding and decoding, ensure that attention is focused on critical areas, enriching both spatial and semantic accuracy [8].

To further enhance the fusion of multi-channel spatial and semantic information, recent work has introduced hypergraph-enhanced learning models, which are particularly effective in capturing relational dependencies across multiple feature spaces. These approaches demonstrate how multi-channel feature integration techniques contribute to more comprehensive and context-aware segmentation, supporting the design of the proposed multi-scale fusion module [9], [10]. By integrating spatial, relational, and contextual information, these methods help improve robustness in scenarios where anatomical boundaries are complex and subtle. Complementary techniques for improving model generalization and adaptation, particularly in scenarios with limited training data, have also contributed to the methodological foundation of this work [11]. These meta-learning principles directly contribute to improving the generalization ability of the segmentation framework.

Addressing class imbalance and sparse data challenges, adaptive weighting approaches have introduced dynamic loss rebalancing strategies that prioritize underrepresented features during training. [12]. These approaches enhance robustness by incorporating realistic synthetic samples into the training set, allowing segmentation models to better handle rare anatomical variations and subtle pathological features [13][14].

Attention-driven feature selection and dynamic feature refinement strategies also benefit from techniques originally applied to entity extraction and hierarchical attention models. These approaches demonstrate how multi-stage attention mechanisms dynamically refine feature importance at different layers, contributing to sharper boundary detection and more robust spatial consistency in segmentation networks [15], [16]. This adaptive focus mechanism directly inspires the attention refinement strategy used in the proposed work.

The exploration of data-driven interface optimization and adaptive human-computer interaction frameworks also contributes indirectly to improving model usability and interpretability. Systematic feature selection methods based on user perception analysis highlight the importance of designing models capable of delivering interpretable outputs aligned with end-user expectations, which supports the visualization and clinical usability components of the proposed model [17]. Similarly, hierarchical control strategies for spatial feature prioritization further emphasize the benefits of multi-level attention when working with complex data hierarchies [18].

Finally, advances in transfer learning applied to medical classification tasks emphasize the importance of pre-trained feature extractors and domain adaptation strategies for improving segmentation performance on limited datasets. These approaches illustrate the benefits of leveraging external domain knowledge to enhance feature transferability, particularly for models trained on limited spinal image datasets [19], [20]. Such strategies contribute to the initialization and fine-tuning processes in the proposed model, ensuring more efficient learning from sparse, domain-specific data.

## III. METHOD

This study presents an improved SwinUNETR-based 3D semantic segmentation algorithm for spine images, aiming to enhance the accuracy and robustness of spine image segmentation. Compared to the traditional SwinUNETR, the innovations of this study lie in two key aspects: First, we introduce a multi-scale fusion mechanism based on contextual information to strengthen the model's ability to learn features at different scales, particularly for capturing fine details in spine images[21]. Second, we have improved the original Swin Transformer architecture by incorporating an adaptive attention mechanism to better focus on important regions within spine images, enhancing the model's ability to extract boundaries and anatomical details, especially in the presence of blurred edges. The model architecture is shown in Figure 1.

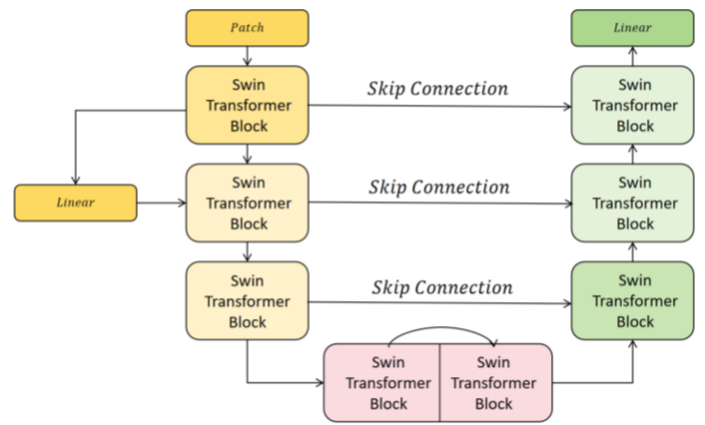

Figure 1 Overall model architecture

The improved method we proposed first focuses on how to extract more comprehensive features from information at different scales. Specifically, for each input three-dimensional image, we first preprocess and standardize it to ensure the uniformity and operability of the input data. Assume that the input three-dimensional image data is $X \in R^{C \times H \times W \times D}$, where C represents the number of channels, and H, W, and D are the height, width, and depth of the image, respectively. We process the input image through a multi-scale convolution module to extract information at different scales. Specifically, we designed a module containing multiple different convolution kernel sizes to process the input image and obtain feature maps $F_1, F_2, ..., F_n$ at different scales, where $F_i \in R^{C_i \times H_i \times W_i \times D_i}$ is the output of the i-th convolution layer. Then, an adaptive fusion module is used to perform weighted fusion on these feature maps of different scales to obtain a multi-scale feature representation $F_{multi-scale}$, whose formula is as follows:

$$F_{multi-scale} = \sum_{i=1}^{n} w_i \cdot F_i$$

Among them, $w_i$ is the weight coefficient, which indicates the importance of each layer of features. The weight coefficient can be obtained through training. This multi-scale feature representation can effectively extract global and local information in spinal images and improve segmentation accuracy.

Based on multi-scale information fusion, the second innovation of the model is the improved adaptive attention mechanism. In order to better capture the key features in spinal images, we introduced an adaptive attention mechanism that can adaptively adjust the attention weights according to different regions of the input image so that the model can pay more attention to the key areas of the spine, especially those with complex anatomical structures and blurred boundaries. Specifically, we used a local attention module based on Swin Transformer, the calculation process of which is as follows:

$$A = Soft\max(\frac{QK^T}{\sqrt{d_k}})$$

Among them, Q and K represent the query and key vectors respectively, $d_k$ is the dimension of the key vector, and A is the calculated attention matrix. By introducing an adaptive attention mechanism, the model can dynamically adjust the attention of each region, so that the details of the spine can receive more attention, thereby improving the accuracy of segmentation.

Combining these two innovations, we built a 3D semantic segmentation model of the spine based on the improved SwinUNETR. The overall structure of the model includes a Swin Transformer encoder, a multi-scale information fusion module, an adaptive attention mechanism module, and a U-Net style decoder. The input 3D image is extracted through the encoder and processed in the multi-scale fusion module. Then, the key areas are enhanced through the adaptive attention mechanism, and finally the segmentation result is output through the decoder.

In order to further improve the training effect of the model, we use a loss function that combines the cross entropy loss function with the Dice coefficient to guide the model training process. Assuming that the predicted output of the model is $Y'$ and the true label is Y, the loss function can be expressed as:

$$L = L_{CE}(Y', Y) + \lambda \cdot L_{Dice}(Y', Y)$$

Among them, $L_{CE}$ is the cross entropy loss function, and the calculation formula is:

$$L_{CE} = -\sum_{i=1}^{N} Y_i \log(Y'_i)$$

$L_{Dice}$ is the Dice coefficient loss function, which is used to measure the overlap between the segmentation result and the true label. The calculation formula is:

$$L_{Dice} = 1 - \frac{2\sum_{i=1}^{N} Y_i Y'_i}{\sum_{i=1}^{N} Y_i + \sum_{i=1}^{N} Y'_i}$$

Here, A is a hyperparameter that controls the weights of the two loss functions. Through this combination of loss functions, we can effectively balance classification accuracy and structure preservation, thereby improving the performance of 3D semantic segmentation of the spine.

In summary, this study has effectively improved SwinUNETR by introducing a multi-scale fusion mechanism and an adaptive attention mechanism, optimizing the accuracy and robustness of spine 3D semantic segmentation. These innovative methods can effectively cope with the complexity and challenges in spine image segmentation, providing a more efficient and accurate solution for medical image analysis.

IV. EXPERIMENT

*A. Datasets*

The dataset used in this study comprises 3D spine images sourced from various high-quality imaging modalities, such as CT and MRI scans. All data collection adhered to strict medical ethics guidelines and encompassed a diverse range of patients, including those of different ages, genders, and pathological conditions. The dataset encompasses the cervical, thoracic, and lumbar spine regions to ensure comprehensive anatomical coverage.

To enhance data diversity and improve the generalizability of the resulting models, the images underwent a series of preprocessing steps, including denoising, contrast enhancement, cropping, and rotation. These steps were performed by professional radiologists who provided pixel-level annotations for key structures, such as vertebral bodies and nerve roots. A multi-round review process was conducted to ensure the accuracy and consistency of the annotations. The dataset size is substantial enough to support effective training, validation, and testing, effectively capturing the intricacies of spine 3D segmentation. An illustrative example is presented in Figure 2.

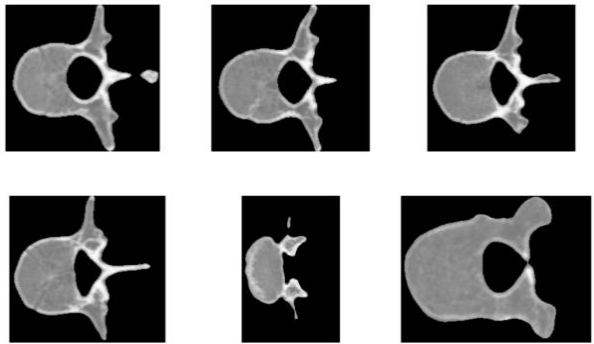

Figure 2. Spine segmentation image

*B. Experimental Results*

In order to verify the effectiveness of the proposed improved SwinUNETR model in the task of 3D semantic segmentation of the spine, this study conducted comparative experiments with several mainstream 3D medical image segmentation methods. Specifically, we selected a method based on a 3D convolutional neural network (3D CNN) as a representative of traditional deep learning models to evaluate its performance in extracting spatial information [23]. In addition, we also selected 3D U-Net as a comparison benchmark. As a classic 3D medical image segmentation network, this model performs well in a variety of medical image segmentation tasks. In order to further analyze the role of the Transformer structure in 3D medical image segmentation, we additionally constructed a version combining 3D U-Net with an ordinary Transformer, and compared its performance with that of the improved SwinUNETR. Through these comparative experiments, we can comprehensively evaluate the segmentation accuracy, boundary processing capabilities, and adaptability of the improved SwinUNETR on complex spinal images, thereby verifying its advantages over traditional methods. The experimental results are shown in Table 1.

Table 1  Experimental Results

| Model | mIou | mDice | mAcc |
|---|---|---|---|
| 3DCNN | 72.4 | 78.9 | 85.2 |
| 3DU-NET | 78.6 | 83.5 | 88.7 |
| 3DU-NET+Transformer | 81.2 | 85.9 | 90.3 |
| Ours | 84.7 | 89.1 | 93.2 |

Experimental results show that the improved SwinUNETR outperforms all baseline models, particularly in mIoU, mDice, and mAcc. Traditional 3D CNN achieved only 72.4% mIoU, 78.9% mDice, and 85.2% mAcc due to limited spatial feature extraction. 3D U-Net improved performance to 78.6% mIoU but struggled with complex boundaries.

The 3D U-Net + Transformer reached 81.2% mIoU, showing better global feature modeling, but still lacked fine detail accuracy. The improved SwinUNETR achieved 84.7% mIoU, 89.1% mDice, and 93.2% mAcc, confirming the effectiveness of the proposed multi-scale fusion and adaptive attention mechanisms. Ablation experiments further verified that both mechanisms significantly enhance segmentation accuracy, demonstrating their necessity for accurate spine 3D segmentation.

Table 2  Results of the Ablation Experiments

| Model | Inference Accuracy | Contextual understanding | Ability to handle complex problems |
|---|---|---|---|
| Ours | 84.7 | 89.1 | 93.2 |
| Remove multi-scale fusion | 81.9 | 86.3 | 90.8 |
| Removing Adaptive Attention | 82.4 | 86.7 | 91.2 |
| Baseline | 79.8 | 84.1 | 88.6 |

The experimental results, presented in Table 2, comprehensively evaluate the proposed models through a series of ablation experiments. The findings reveal that both the multi-scale fusion mechanism and the adaptive attention mechanism significantly contribute to the segmentation performance of the model. Notably, after removing multi-scale fusion, mIoU, mDice, and mAcc decreased by 2.8%, 2.8%, and 2.4%, respectively. This indicates that multi-scale information is crucial for capturing diverse structural features of spinal images. Similarly, removing adaptive attention also led to a decrease in the model's performance, particularly in mDice and mAcc, highlighting the importance of the adaptive attention mechanism in enhancing feature expression and improving segmentation accuracy. Moreover, the baseline model, which lacked both modules, performed the worst in all indicators. Notably, mIoU decreased by 4.9% compared to the complete model, further validating the effectiveness of the proposed method. These results demonstrate that the enhanced method of this study significantly improves the model's ability to comprehend spinal 3D images, enhances segmentation accuracy and stability, and makes it more suitable for real-world medical image analysis tasks.

Finally, the visualization result is given, as shown in Figure 3.

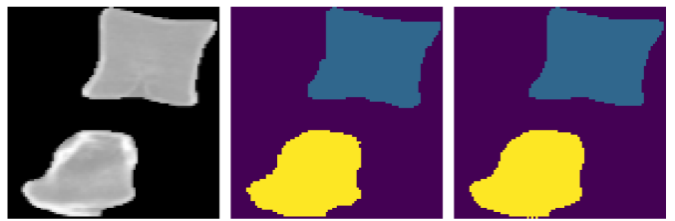

Figure 3. Single sample inference segmentation results

From the visualization results in Figure 3, we can see that the model can predict the target area well in the 3D semantic segmentation task of the spine. The first picture is the input original medical image. The second picture is the corresponding Ground Truth (real label), in which different colors represent different anatomical structures, showing the standard segmentation area that the model should learn. The third picture is the inference result of the model. Compared with the real label, it can be found that the model can accurately predict the boundary of the target area, and the main structure is basically consistent with the Ground Truth, indicating that the model has good performance in the segmentation task. Although the overall segmentation effect is relatively accurate, there are still some subtle errors. For example, some edge areas in the inference results are slightly deviated, which may be due to the ambiguity of the model when processing complex boundary information. In addition, the color filling of some areas is slightly incomplete, which may indicate that the model has a certain degree of misjudgment of some small targets or areas with blurred boundaries. This may be due to the limitation of the amount of training data or the model's neglect of some information during the feature extraction process.

## V. Conclusion

This study proposed a 3D semantic segmentation method for the spine based on the improved SwinUNETR. By introducing a multi-scale fusion mechanism and an adaptive attention mechanism, the model's feature extraction capability for spine images was enhanced and the segmentation accuracy was improved. Experimental results show that compared with 3D CNN, 3D U-Net, and 3D U-Net + Transformer, the model of this study has achieved significant improvements in mIoU, mDice, and mAcc, proving the effectiveness of the proposed method. At the same time, the ablation experiment further verified the important role of multi-scale fusion and adaptive attention in improving the performance of the model, indicating that these improvements can effectively improve the segmentation performance of the model under complex anatomical structures. In addition, the qualitative analysis results also show that the improved SwinUNETR can better capture the detailed information of the target area, making the segmentation results closer to the real annotation.

Although the model of this study has achieved good performance in the 3D semantic segmentation task of the spine, there is still room for optimization. Future research can further explore more efficient Transformer structures to reduce computational overhead and increase inference speed. In addition, at the data level, the generalization ability of the model can be enhanced by introducing more diverse spinal image data so that it can maintain stable performance on different types of image data. Overall, the method of this study provides an effective solution for spinal medical image analysis and provides a new research direction for future medical image segmentation tasks.

## References

[1] H. Zhao, W. Zhu, X. Deng, et al., "Automatic 2D/3D spine registration based on two-step transformer with semantic attention and adaptive multi-dimensional loss function," Biomedical Signal Processing and Control, vol. 95, no. 106384, 2024.

[2] R. Pal, S. Mondal, A. Gupta, et al., "Lumbar spine tumor segmentation and localization in T2 MRI images using AI," arXiv preprint arXiv:2405.04023, 2024.

[3] S. Gu, J. K. Adhinarta, M. Bessmeltsev, et al., "FreSeg: Frenet-frame-based part segmentation for 3D curvilinear structures," arXiv preprint arXiv:2404.14435, 2024.

[4] K. Ibragimov, G. Podobnik, T. Trojner, T. N. Tuleasca, F. N. Fotiadou and B. Likar, "Mid-sagittal cross-section identification for vertebra landmarking in MR spine images," Proceedings of Medical Imaging 2024: Image Processing, vol. 12926, pp. 776-781, 2024.

[5] X. Li, Q. Lu, Y. Li, M. Li and Y. Qi, "Optimized Unet with attention mechanism for multi-scale semantic segmentation," arXiv preprint arXiv:2502.03813, 2025.

[6] Y. Qi, Q. Lu, S. Dou, X. Sun, M. Li and Y. Li, "Graph neural network-driven hierarchical mining for complex imbalanced data," arXiv preprint arXiv:2502.03803, 2025.

[7] J. Gao, G. Liu, B. Zhu, S. Zhou, H. Zheng and X. Liao, "Multi-level attention and contrastive learning for enhanced text classification with an optimized transformer," arXiv preprint arXiv:2501.13467, 2025.

[8] L. Wu, J. Gao, X. Liao, H. Zheng, J. Hu and R. Bao, "Adaptive attention and feature embedding for enhanced entity extraction using an improved BERT model," unpublished.

[9] Z. Gao, T. Mei, Z. Zheng, X. Cheng, Q. Wang and W. Yang, "Multi-channel hypergraph-enhanced sequential visit prediction," Proceedings of the 2024 International Conference on Electronics and Devices, Computational Science (ICEDCS), pp. 421-425, 2024.

[10] T. Mei, Z. Zheng, Z. Gao, Q. Wang, X. Cheng and W. Yang, "Collaborative hypergraph networks for enhanced disease risk assessment," Proceedings of the 2024 International Conference on Electronics and Devices, Computational Science (ICEDCS), pp. 416-420, 2024.

[11] J. Gao, S. Lyu, G. Liu, B. Zhu, H. Zheng and X. Liao, "A hybrid model for few-shot text classification using transfer and meta-learning," arXiv preprint arXiv:2502.09086, 2025.

[12] J. Wang, "Markov network classification for imbalanced data with adaptive weighting," Journal of Computer Science and Software Applications, vol. 5, no. 1, pp. 43-52, 2025.

[13] X. Wang, "Data mining framework leveraging stable diffusion: a unified approach for classification and anomaly detection," Journal of Computer Technology and Software, vol. 4, no. 1, 2025.

[14] X. Liao, B. Zhu, J. He, G. Liu, H. Zheng and J. Gao, "A fine-tuning approach for T5 using knowledge graphs to address complex tasks," arXiv preprint arXiv:2502.16484, 2025.

[15] Z. Zhu, Y. Zhang, J. Yuan, W. Yang, L. Wu and Z. Chen, "NLP-driven privacy solutions for medical records using transformer architecture," unpublished.

[16] J. Wang, "Multivariate time series forecasting and classification via GNN and transformer models," Journal of Computer Technology and Software, vol. 3, no. 9, 2024.

[17] S. Duan, "Systematic analysis of user perception for interface design enhancement," Journal of Computer Science and Software Applications, vol. 5, no. 2, 2024.

[18] Q. Sun, "Spatial hierarchical voice control for human-computer interaction: performance and challenges," Journal of Computer Technology and Software, vol. 4, no. 1, 2025.

[19] X. Yan, et al., "Survival prediction across diverse cancer types using neural networks," Proc. ICVA, 2024.

[20] W. Wang, et al., "Breast cancer image classification using deep transfer learning," Proc. IPMLPR, 2024.

[21] Y. Rong, L. Nong, Z. Liang, et al., "Hypergraph position attention convolution networks for 3D point cloud segmentation," Applied Sciences, vol. 14, no. 8, pp. 3526, 2024.

[22] R. Zhou, Y. Feng, G. Wang, et al., "TSUBF-Net: Trans-spatial UNet-like network with bi-direction fusion for segmentation of adenoid hypertrophy in CT," Neural Computing and Applications, pp. 1-17, 2025.

[23] B. K. Bhagyalaxmi and B. Dwarakanath, "CDCG-UNet: Chaotic optimization assisted brain tumor segmentation based on dilated channel gate attention U-Net model," Neuroinformatics, vol. 23, no. 2, pp. 1-26, 2025.